\documentclass[letterpaper, 10 pt, journal, twoside]{IEEEtran}

\usepackage{myStyle}

\begin{document}



\title{Multi-Modal MPPI and Active Inference for Reactive Task and Motion Planning}

\author{Yuezhe Zhang, Corrado Pezzato, Elia Trevisan, Chadi Salmi, Carlos Hern\'andez Corbato, Javier Alonso-Mora


\thanks{This research was supported in part by Ahold Delhaize; by the Netherlands Organization for Scientific Research (NWO), domain Science (ENW), TRILOGY project; and by the European Union through ERC, under Grant 101041863 (INTERACT). \textit{(Corresponding author: Yuezhe Zhang.)}}
\thanks{The authors are with Cognitive Robotics Department,
        TU Delft, The Netherlands
        {\tt\small {yuezhezhang\_bit@163.com}, \{corrado.pezzato, salmi.chadi\}@gmail.com, \{e.trevisan, c.h.corbato, j.alonsomora\}@tudelft.nl}}

}

\markboth{IEEE Robotics and Automation Letters. PREPRINT VERSION.}
{Zhang \MakeLowercase{\textit{et al.}}: Multi-Modal MPPI and Active Inference for Reactive Task and Motion Planning} 

\maketitle

\IEEEpeerreviewmaketitle
\begin{abstract}
Task and Motion Planning (TAMP) has made strides in complex manipulation tasks, yet the execution robustness of the planned solutions remains overlooked. In this work, we propose a method for reactive TAMP to cope with runtime uncertainties and disturbances. We combine an Active Inference planner (AIP) for adaptive high-level action selection and a novel Multi-Modal Model Predictive Path Integral controller (M3P2I) for low-level control. This results in a scheme that simultaneously adapts both high-level actions and low-level motions. The AIP generates alternative symbolic plans, each linked to a cost function for M3P2I. The latter employs a physics simulator for diverse trajectory rollouts, deriving optimal control by weighing the different samples according to their cost. This idea enables blending different robot skills for fluid and reactive plan execution, accommodating plan adjustments at both the high and low levels to cope, for instance, with dynamic obstacles or disturbances that invalidate the current plan. We \textcolor{black}{have tested} our approach in simulations and real-world scenarios. \textcolor{blue}{Project website:} \href{https://autonomousrobots.nl/paper_websites/m3p2i-aip}{https://autonomousrobots.nl/paper\_websites/m3p2i-aip}

\end{abstract}

\section{Introduction}
\label{sec:intro}
\IEEEPARstart{T}{ask} and Motion Planning (TAMP) is a powerful class of methods for solving complex long-term manipulation problems where logic and geometric variables influence each other. TAMP \cite{garrett2021integrated, toussaint2015logic, garrett2020pddlstream} has been successfully applied to domains such as table rearrangement, stacking blocks, or solving the Hanoi tower. \textcolor{blue}{However, the plan is often executed in open-loop in static environments.} Recent works \cite{toussaint2022sequence, migimatsu2020object,li2021reactive} recognized the importance of robustifying the execution of TAMP plans to be able to carry them out in the real world reliably. But they either rely only on the adaptation of the action sequence in a plan \cite{pezzato2022aipbt, castaman2021receding, Colledanchise2019, li2021reactive} or only on the motion planning problem in a dynamic environment given a fixed plan \cite{toussaint2022sequence, migimatsu2020object}. \textcolor{blue}{Unlike typical TAMP planners that focus on solving static and complex tasks offline and then execute the solution, this paper aims to achieve reactive execution by simultaneously adapting high-level actions and low-level motions}. 

\textcolor{blue}{Reactive TAMP faces the challenge of accommodating unforeseen geometric constraints during planning, such as the need to pull rather than push a block when it's in a corner, complicating high-level planning without complete scene knowledge. Additionally, scenarios like pick-and-place tasks with dynamic obstacles and human disturbances demand varied grasping poses for different objects and obstacles, requiring TAMP algorithms to adapt to such configurations dynamically.}

We address these challenges by proposing a control scheme that jointly achieves reactive action selection and robust low-level motion planning during execution. We propose a high-level planner capable of providing alternative actions to achieve a goal. These actions are translated to different cost functions for our new Multi-Modal Model Predictive Path Integral controller for motion planning. \textcolor{black}{This} motion planner leverages a physics simulator to sample parallel motion plans that minimize the given costs and computes one coherent control input that effectively blends different strategies. To achieve this, we build upon two of our recent works: 1) an Active Inference planner (AIP) \cite{pezzato2022aipbt} for symbolic action selection, and 2) a Model Predictive Path Integral (MPPI) controller \cite{pezzato2023sampling} for motion planning. \textcolor{blue}{The AIP computes a sequence of actions and state transitions through backchaining to achieve a sub-goal specified in a given Behavior Tree (BT). The BT guides the search and allows real-time high-level planning within the AIP framework \cite{pezzato2022aipbt}. In this work,} we extend the previous AIP to plan possible alternative \textcolor{blue}{action} plans, and we propose a new Multi-Modal Model Predictive Path Integral controller (M3P2I) that can sample in parallel these alternatives and smoothly blend them \textcolor{blue}{considering the geometric constraints of the problem}.

\subsection{Related work}



\textcolor{black}{To robustly operate in dynamic environments, reactive motion planners are necessary.} In \cite{toussaint2022sequence}, the authors provided a reactive Model Predictive Control (MPC) strategy to execute a TAMP plan as a given linear sequence of constraints. The reactive nature of the approach allows coping with disturbances and dynamic collision avoidance during the execution of a TAMP plan. Authors in \cite{migimatsu2020object} formulated a TAMP plan in object-centric Cartesian coordinates, showing how this allows coping with perturbations such as moving a target location. However, both \cite{toussaint2022sequence, migimatsu2020object} do not consider adaptation at the symbolic action level if a perturbation invalidates the current plan. 

Several papers focused on adapting and repairing high-level action sequences during execution.  In \cite{Paxton2019}, robot task plans are represented as robust logical-dynamical systems to handle human disturbances. Similarly, \cite{harris2022fc} coordinates control chains for robust plan execution through plan switching and controller selection. 
A recent paper \cite{huang2022parallel} suggests employing Monte Carlo Tree Search with IsaacGym to accelerate task planning for multi-step object retrieval from clutter involving intricate physical interactions. While promising, \cite{huang2022parallel} only supports high-level reasoning with predefined motions in an open loop. Recent works~\cite{zhou22, li2021reactive} combined BTs and linear temporal logic to adapt the high-level plan to cope with cooperative or adversarial human operators, environmental changes, or failures. In our previous work \cite{pezzato2022aipbt}, AIP and BT were combined to provide reactive action selection in long-term tasks in partially observable and dynamic environments. This method achieved hierarchical deliberation and continual online planning, making it particularly appealing for the problem of reactive TAMP at hand. In this paper, we extend \cite{pezzato2022aipbt} by bridging the gap to low-level reactive control by planning cost functions instead of symbolic actions.  

At the lower level, MPC is a widely used approach \cite{bangura2014real, scianca2020mpc, spahn2021coupled}. However, manipulation tasks often involve discontinuous contacts that are hard to differentiate. Sampling-based MPCs, such as MPPI \cite{williams2017information,bhardwaj2022storm}, can handle non-linearities, non-convexities, or discontinuities of the dynamics and costs. MPPI relies on sampling control input sequences and forward system dynamics simulation. The resulting trajectories are weighted according to their cost to approximate an optimal control input. In \cite{abraham2020model}, the authors proposed an ensemble MPPI to cope with model parameters uncertainty. Sampling-based MPCs are generally applied for single-skill execution, such as pushing or reaching a target point. As pointed out in the future work of \cite{howell2022predictive}, one could use a high-level agent to set the cost functions for the sampling-based MPC for long-horizon cognitive tasks. We follow this line of thought and propose a method to reactively compose cost functions for long-horizon tasks. Moreover, classical MPPI approaches can only keep track of one cost function at a time. 
This means the task planner should propose a single plan to solve the task. However, some tasks might present geometric ambiguities for which multiple plans could be effective, and selecting what strategy to pursue can only be determined by the motion planner based on the geometry of the problem. 

\subsection{Contributions}
The main contribution of this work is a reactive task and motion planning algorithm based on the following: 
\begin{itemize}
    \item A new Multi-Modal MPPI (M3P2I) capable of sampling in parallel plan alternatives to achieve a goal, evaluating them against different costs. This enables the smooth blending of alternative solutions into a coherent behavior instead of switching based on heuristics.
    \item An enhanced Active Inference planner (AIP) capable of generating alternative cost functions for M3P2I.
\end{itemize}

We demonstrate the method in several scenarios in simulations and real robots for pushing, pulling, picking, and placing objects under disturbances.

\section{Background}
\label{sec:background}
In this section, we present the background knowledge about the Active Inference planner and Model Predictive Path Integral Control to understand the contributions of this paper. We refer the interested reader to the original articles \cite{pezzato2022aipbt,williams_model_2017, williams_information-theoretic_2018} for a more in-depth understanding of the techniques. 
\subsection{Active Inference Planner (AIP)}
AIP is a high-level decision-making algorithm that relies on symbolic states, observations, and actions \cite{pezzato2022aipbt}. Each independent set of states in AIP is a factor, and the planner contains a total of $n_f$ factors. For a generic factor $f_j$ where $j\in\mathcal{J} = \{1,...,n_f\}$, it holds: 
\begin{gather}
    \nonumber
    s^{(f_j)} = \left[s^{(f_j,1)}, s^{(f_j,2)},...,s^{(f_j,m^{(f_j)})}\right]^\top,\\ 
    \mathcal{S} = \big\{ s^{(f_j)}|j\in\mathcal{J}\big\}
\end{gather}
where $m^{(f_j)}$ is the number of mutually exclusive symbolic values a state factor can have, each entry of $s^{(f_j)}$ is a real value between 0 and 1, and the sum of the entries is 1. This represents the current belief state.

The continuous state of the world $x\in\mathcal{X}$ is discretized through a symbolic observer such that the AIP can use it. Discretized observations $o$ are used to build a probabilistic belief about the symbolic current state. Assuming one set of observations per state factor with $r^{(f_j)}$ possible values:
\begin{gather}
    \nonumber
    o^{(f_j)} = \left[o^{(f_j,1)}, o^{(f_j,2)},...,o^{(f_j,r^{(f_j)})}\right]^\top,\\ 
    \mathcal{O} = \big\{ o^{(f_j)}|j\in\mathcal{J}\big\}
\end{gather}

The robot has a set of symbolic actions that can act then their corresponding state factor:
\begin{gather}
    \nonumber
    a_\tau \in \alpha^{(f_j)} = \big\{a^{(f_j,1)}, a^{(f_j,2)},...,a^{(f_j,k^{(f_j)})}\big\},\\ 
    \mathcal{A} = \big\{ \alpha^{(f_j)}|j\in\mathcal{J}\big\}
\end{gather}
where $k^{(f_j)}$ is the number of actions that can affect a specific state factor $f_j$. Each generic action $a^{(f_j,\cdot)}$ has associated a symbolic name, \textit{parameters}, \textit{pre-} and \textit{postconditions}:
\begin{table}[ht]
\centering 
\begin{tabular}{l c c} 
    \textbf{Action} $a^{(f_j,\cdot)}$& \textbf{Preconditions}& \textbf{Postconditions}\\  %
    \texttt{action\_name(}$par$\texttt{)}&\texttt{prec}$_{a^{(f_j,\cdot)}}$ & \texttt{post}$_{a^{(f_j,\cdot)}}$
\end{tabular}
\end{table}

\noindent where \texttt{prec}$_{a^{(f_j,\cdot)}}$ and \texttt{post}$_{a^{(f_j,\cdot)}}$ are \textit{first-order logic predicates} that can be evaluated at run-time. A logical predicate is a boolean-valued function $\mathcal{B}:\mathcal{X}\rightarrow\{$\texttt{true}, \texttt{false}$\}$. Finally, we define the logical state $l^{(f_j)}$ as a one-hot encoding of $s^{(f_j)}$. The AIP computes the posterior distribution over $p$ plans $\bm \pi$ through free-energy minimization \cite{pezzato2022aipbt}. The symbolic action to be executed by a robot in the next time step is the first action of the most likely plan, denoted with $\pi_{\zeta, 0}$:
\begin{eqnarray}
\label{eq:a_t}
    \zeta = \max(\underbrace{[\bm\pi_{1}, \bm\pi_{2},...,\bm\pi_{p}]}_{\bm \pi^\top}),\ 
    a_{\tau=0} = \pi_{\zeta, 0}.
\end{eqnarray}
\subsection{Model Predictive Path Integral Control (MPPI)}
MPPI is a method for solving optimal stochastic problems in a sampling-based fashion~\cite{williams_model_2017, williams_information-theoretic_2018}. Let us consider the following discrete-time systems:
\begin{equation}
    x_{t+1} = f(x_t, v_t),\ \ \ v_t \sim \mathcal{N}(u_t, \Sigma),
\end{equation}
where $f$, a nonlinear state-transition function, describes how the state $x$ evolves over time $t$ with a control input $v_t$. $u_t$ and $\Sigma$ are the commanded input and the variance, respectively. $K$ noisy input sequences ${V}_k$ are sampled and then applied to the system to forward simulate $K$ state trajectories $Q_k$, $k \in [0,K-1]$, over a time horizon $T$. Given the state trajectories $Q_k$ and a designed cost function $C$ to be minimized, the total state-cost $S_k$ of an input sequence $V_k$ is computed by evaluating $S_k = C(Q_k)$. Finally, each rollout is weighted by the importance sampling weights $w_k$. These are computed through an inverse exponential of the cost $S_k$ with tuning parameter $\beta$ and normalized by $\eta$. For numerical stability, the minimum sampled cost $\rho = \min_k S_k$ is subtracted, leading to:
\begin{equation}
    \label{eq:weights}
    w_k = \frac{1}{\eta}\exp\left(-\frac{1}{\beta}(S_k - \rho)\right), \ \ \ \sum_{k=1}^K w_k=1
\end{equation}

The parameter $\beta$ is called \textit{inverse temperature}. The importance sampling weights are finally used to approximate the optimal control input sequence $U^*$:
\begin{equation}
    \label{eq:approx_U}
    U^* = \sum_{k=1}^K w_k {V}_k
\end{equation}

The first input $u_0^*$ of the sequence $U^*$ is applied to the system, and the process is repeated. At the next iteration, $U^*$ is used as a warm-start, time-shifted backward of one timestep. Specifically, the second last input in the shifted sequence is also propagated to the last input. In this work, we build upon our previous MPPI approaches \cite{pezzato2023sampling, trevisan2024biased}, where we employed IsaacGym as a dynamic model to forward simulate trajectory rollouts \textcolor{blue}{and allow for arbitrary sampling distributions}. 

\section{Methodology}

\label{sec:algorithm}
\begin{figure}[!b]
    \centering
    \includegraphics[width=0.42\textwidth]{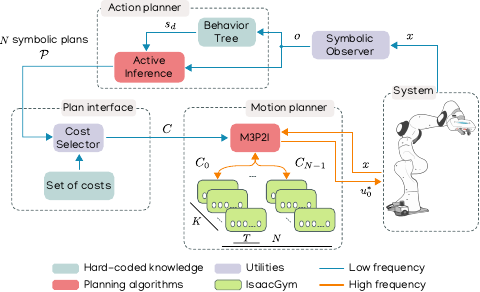}
    \caption{Proposed scheme. Given symbolic observations $o$ of the environment, the action planner computes $N$ different plan alternatives linked to individual cost functions $C_i$. M3P2I samples control input sequences and uses an importance sampling scheme to approximate the optimal control $u_0^*$.}
    \label{fig:scheme}
\end{figure}

The proposed method is depicted in \Cref{fig:scheme}. After a general overview, we discuss the three main parts of the scheme:  \textit{action planner}, \textit{motion planner}, and \textit{plan interface}.



\subsection{Overview}
The proposed scheme works as follows. First, the \textit{symbolic observers} translates continuous states $x$ into discretized symbolic observations $o$\textcolor{blue}{, which are then passed to the action planner. The current desired state $s_d$ for Active Inference can be manually set or be encoded as the skeleton solution of a BT as previous work \cite{pezzato2022aipbt}}. The \textcolor{blue}{AIP} computes $N$ alternative symbolic plans based on the current symbolic state and the available symbolic actions. The symbolic actions are encoded as action templates with pre-post conditions that Active Inference uses to construct action sequences to achieve the desired state. After the plans are generated, the plan interface links the first action $a_{0,i},\ i=0...N-1$ of each plan to a cost function $C_i$. The cost functions are sent to {M3P2I}, which samples $N\cdot K$ different control input sequences. The input sequences are forward simulated using IsaacGym, which encodes the dynamics of the problem~\cite{pezzato2023sampling}. The resulting trajectories are evaluated against their respective costs. Finally, an importance sampling scheme calculates the approximate optimal control $u_0^*$. All processes are running continuously during execution at different frequencies. The action planner runs, for instance, at $1Hz$ while the motion planner runs at $25Hz$. An overview can be found in \Cref{alg:whole scheme}. 
\begin{algorithm}[htb!]
    \caption{Overview of the method}\label{alg:whole scheme}
    \begin{algorithmic}[1]
        \STATE \textcolor{blue}{\textbf{Input:} action templates and inputs from \Cref{alg:alternative plans} to \ref{alg:m3p2i}}
        \STATE \textcolor{blue}{$AIP.task = AIP.agent(ActionTemplates)$ }
        \WHILE{\textcolor{blue}{t}ask not completed}
            \STATE $o \leftarrow GetSymbolicObservation(x)$
            \STATE /* \textit{Get current desired state} */
            \STATE $AIP.s_d \leftarrow BT(o)$ \textcolor{blue}{or be manually set}\hfill $\triangleright$ from \cite{pezzato2022aipbt}
            \STATE /* \textit{Get current action plans from Active Inference} */
            \STATE $\mathcal{P} \leftarrow AIP.par\textcolor{blue}{a}ll\_act\_sel(o) $ \hfill $\triangleright$ \Cref{alg:alternative plans}
            \STATE /* \textit{Translate action plan to cost function} */
            \STATE $C \leftarrow Interface(\mathcal{P})$
            \STATE /* \textit{Compute motion commands} */
            \STATE $M3P2I.command(C)$ \hfill $\triangleright$ \Cref{alg:m3p2i}
        \ENDWHILE
    \end{algorithmic}
\end{algorithm}


\subsection{Action planner - Active Inference Planner (AIP)}
In contrast to our previous work~\cite{pezzato2022aipbt} where only one action $a_\tau$ for the next time step is computed, we modify the AIP to generate action alternatives. In particular, instead of stopping the search for a plan when a valid executable action $a_\tau$ is found, we repeat the search while removing that same $a_\tau$ from the available action set $\mathcal{A}$. This simple change is effective because we are looking for alternative actions to be applied at the next step, and the AIP builds plans backward from the desired state \cite{pezzato2022aipbt}. The pseudocode is reported in \Cref{alg:alternative plans}. The algorithm will cease when no new actions are found, returning a list of possible plans $\mathcal{P}$. This planner is later integrated with M3P2I to evaluate different alternatives in real-time. This increases the \textcolor{blue}{robustness at run-time and,} at the same time, reduces the number of heuristics to be encoded in the action planner. \textcolor{blue}{Specifically, one does not need to encode when to prefer a symbolic action over another based on the geometry of the problem}.

\begin{algorithm}[htb!]
    \caption{Generate alternative plans using Active Inference}\label{alg:alternative plans}
    \begin{algorithmic}[1]
        \STATE \textcolor{blue}{\textbf{Input:} available action set: $\mathcal{A}$ } 
        \STATE $a_\tau \leftarrow AIP.act\_sel(o)$ \hfill $\triangleright$  from \cite{pezzato2022aipbt}
        \STATE Set $\mathcal{P} \leftarrow \emptyset$
        \WHILE{$a_\tau !=$ \texttt{none}}
            \STATE $\mathcal{P}.append(a_\tau)$
            \STATE $\mathcal{A} = \mathcal{A} \symbol{92} \{a_\tau\} $
            \STATE $a_\tau \leftarrow AIP.act\_sel(o)$ \hfill $\triangleright$  from \cite{pezzato2022aipbt}
        \ENDWHILE
        \STATE \textbf{Return} $\mathcal{P}$
    \end{algorithmic}
\end{algorithm}


\subsection{Motion planner - Multi-Modal MPPI (M3P2I)}
We propose a Multi-Modal MPPI capable of sampling different plan alternatives from the AIP. Traditional MPPI approaches consider \textit{one} cost function and \textit{one} sampling distribution.
In this work, we propose keeping track of $N$ separate control input sequences corresponding to $N$ different plan alternatives/costs. This is advantageous because it offers a general approach to exploring different strategies in parallel. We perform $N$ separate sets of importance weights, one for each alternative, and only ultimately, we combine the weighted control inputs in one coherent control. This allows the smooth blending of different strategies. Assume we consider $N$ alternative plans, a total of $N\cdot K$ samples. Assume the cost of plan $i, i \in [0, N)$ to be formulated as:
\begin{equation}
    \label{eq: cost function 1}
    S_i(V_k) = \sum_{t=0}^{T-1}\gamma^t C_i(x_{t, k}, v_{t, k})
\end{equation}
$ \forall k \in \kappa(i)$ where $\kappa(i)$ is the integer set of indexes ranging from $i \cdot K$ to $(i+1)\cdot K-1$. State $x_{t, k}$ and control input $v_{t, k}$ are indexed based on the time $t$ and trajectory $k$. The random control sequence $V_k = [v_{0, k}, v_{1, k}, \dots, v_{T-1, k}]$ defines the control inputs for trajectory $k$ over a time horizon $T$. The trajectory $Q_i(V_k) = [x_{0, k}, x_{1, k}, \dots, x_{T-1, k}]$ is determined by the control sequence $V_k$ and the initial state $x_{0,k}$. $C_i$ is the cost function for plan $i$. Finally, $\gamma \in [0, 1]$ is a discount factor \textcolor{black}{that evaluates the importance of accumulated future costs}. As in classical MPPI approaches, given the costs $S_i(V_k)$, we can compute the importance sampling weights associated with each alternative as:
\begin{align}
    \label{eq: weight 1}
    \omega_i(V_k) &= \frac{1}{\eta_i} \exp \left(
    -\frac{1}{\beta_i} 
    \left( S_i(V_k) - \rho_i \right)
    \right), \forall k \in \kappa(i)\\
    \label{eq:eta1}
    \eta_i &=  \sum_{k \in \kappa(i)} \exp \left(-\frac{1}{\beta_i}
    \left( S_i(V_k) - \rho_i \right)
    \right)\\
    \label{eq:ro1}
    \rho_i &= \min_{k \in \kappa(i)} S_i(V_k)
\end{align}

We use the insight in \cite{pezzato2023sampling} to 1) sample Halton splines instead of Gaussian noise for smoother behavior, 2) automatically tune the inverse temperature $\beta_i$ to maintain the normalization factor $\eta_i$ within certain bounds. The latter is helpful since $\eta_i$ indicates the number of samples to which significant weights are assigned. If $\eta_i$ is close to the number of samples $K$, an unweighted average of sampled trajectories will be taken. If $\eta_i$ is close to 1, then the best trajectory sample will be taken. We observed that setting $\eta_i$ \textcolor{black}{between 5\% and 10\% of }$K$ generates smooth trajectories. As opposed to \cite{pezzato2023sampling}, we update $\eta$ \textit{within a rollout} to stay within bounds instead of updating it once per iteration, see \Cref{alg:update temperature}.
\begin{algorithm}[htb!]
    \caption{Update inverse temperature $\beta_i$}\label{alg:update temperature}
    \begin{algorithmic}[1]
        \STATE \textcolor{blue}{\textbf{Input:} parameters: $\eta_{l}, \eta_{u}$}
        \WHILE{$\eta_i \notin [\eta_{l}, \eta_{u}]$}
        \STATE $\rho_i \leftarrow \min_{k \in \kappa(i)} S_i(V_k)$ \hfill $\triangleright$ \cref{eq:ro1}
            \STATE $\eta_i \leftarrow \sum_{k\in \kappa(i)} \exp (-\frac{S_i(V_k)-\rho_i}{\beta_i})$ \hfill $\triangleright$ \cref{eq:eta1}
            \IF{$\eta_i > \eta_u$} \hfill $\triangleright$ \textcolor{blue}{greater} than upper bound
                \STATE $\beta_i = 0.9 * \beta_i$
            \ELSIF{$\eta_i < \eta_l$} \hfill $\triangleright$ smaller than lower bound
                \STATE $\beta_i = 1.2 * \beta_i$
            \ENDIF
        \ENDWHILE
        \STATE \textbf{Return} $\rho_i, \eta_i, \beta_i$
    \end{algorithmic}
\end{algorithm}
We use $\mu_i$ to denote the action sequence of plan $i$ over a time horizon $\mu_i = [\mu_{i, 0}, \mu_{i, 1}, \dots, \mu_{i, T-1}]$. Each sequence is weighted by the corresponding weights leading to:
\begin{align}
    \label{eq:mean action i}
    \mu_i = \sum_{k \in \kappa(i)} \omega_i(V_k) V_k
\end{align}

At every iteration, we add to $\mu_i$ the sampled noise from \textit{Halton splines} \cite{bhardwaj2022storm}. Then, we forward simulate the state trajectories $Q_i(V_k)$ using IsaacGym as in \cite{pezzato2023sampling}. Finally, given the state trajectories corresponding to the plan alternatives, we need to compute the weights and mean for the overall control sequence. To do so, we concatenate the $N$ state-costs $S_i(V_k), i \in [0, N)$ and represent it as $\tilde{S}(V)$. 
Therefore, we calculate the weights for the whole control sequence as \cite{bhardwaj2022storm}:
\begin{equation}
    \label{eq: weight all}
    \tilde{\omega}(V) = \frac{1}{\eta} \exp \left(
    -\frac{1}{\beta} 
    \left( \tilde{S}(V) - \rho \right)
    \right)
\end{equation}

Similarly, $\eta, \rho$ are computed as in \Cref{eq:eta1,eq:ro1} but considering $\tilde{S}(V)$ instead. 
The overall mean action over time horizon $T$ is denoted as $u = [u_0, u_1, \dots, u_{T-1}]$. For each timestep $t$: 
\begin{align}
    \label{eq:mean action all}
    u_t = (1-\alpha_u)u_{t-1} + \alpha_u\sum_{k = 0}^{N\cdot K-1} \tilde{\omega}_k(V) v_{t,k}
\end{align}
where $\alpha_u$ is the step size that regularizes the current solution to be close to the previous $u_{t-1}$. The optimal control is set to $u_0^* = u_0$. Note that through \Cref{eq: weight all}, we can smoothly fuse different strategies to achieve a goal in a general way.  

\begin{algorithm}[htb!]
\caption{Multi-Modal Model Predictive Path Integral Control (M3P2I)}\label{alg:m3p2i}
\begin{algorithmic}[1]
\STATE \textcolor{blue}{\textbf{Input:} cost functions: $C_i, \forall i \in [0, N)$} 
\STATE Parameters: $N, K, T$
\STATE \textcolor{blue}{Initial sequence:} $\mu_i = \bm 0, u= \bm 0, \in \mathbb{R}^T\ \forall i \in [0, N)$
\WHILE{task not completed}
    \STATE $x \leftarrow GetStateEstimate()$
    \STATE $InitIsaacGym(x)$
    \STATE /* \textit{Begin parallel sampling of alternatives} */
    \FOR{$i=0$ to $N-1$} \label{line:plan}
        \FOR{$k \in \kappa(i)$} \label{line:control}
            \STATE $S_i(V_k) \leftarrow 0$
            \STATE Sample noise $ \mathcal{E}_k \leftarrow SampleHaltonSplines()$
            \STATE $\mu_i \leftarrow BackShift(\mu_i)$
            \FOR{$t=0$ to $T-1$} \label{line:time}
                \STATE $Q_i(V_k) \leftarrow ComputeTrajIsaacGym(\mu_i+\mathcal{E}_k)$  
                \STATE $S_i(V_k) \leftarrow UpdateCost(C_i, Q_i(V_k))$ \hfill $\triangleright$ \cref{eq: cost function 1}
            \ENDFOR
        \ENDFOR
    \ENDFOR \label{line:end_plan}
    \STATE /* \textit{Begin computing trajectory weights} */
    \FOR{$i=0$ to $N-1$} \label{line:weights 0}
        \STATE $\rho_i, \eta_i, \beta_i \leftarrow UpdateInvTemp(i)$ \hfill $\triangleright$ \Cref{alg:update temperature}
        \STATE $\omega_i(k) \leftarrow \frac{1}{\eta_i} \exp (-\frac{1}{\beta_i}(S_i(V_k)-\rho_i)), \forall k$ \hfill $\triangleright$ \cref{eq: weight 1}
        \STATE $\mu_i = \sum_{k \in \kappa(i)} \omega_i(V_k) V_k$ \hfill $\triangleright$ \cref{eq:mean action i}
    \ENDFOR \label{line:weights 1}
    \STATE /* \textit{Begin control update} */
    \STATE $ \tilde{\omega}(V) = \frac{1}{\eta} \exp \left(
    -\frac{1}{\beta} 
    \left( \tilde{S}(V) - \rho \right)
    \right)$ \hfill $\triangleright$ \cref{eq: weight all}
    \FOR{$t=0$ to $T-1$}
        \STATE $u_t = (1-\alpha_u)u_{t-1} + \alpha_u\sum_{j = 0}^{N\cdot K-1} \tilde{\omega}_k v_{t, k}$  \hfill $\triangleright$ \cref{eq:mean action all}\label{line:mean} 
    \ENDFOR
    \STATE $ExecuteCommand(u_0^* = u_0)$ \label{line:execute}
    \STATE $u =  BackShift(u)$
\ENDWHILE
\end{algorithmic}
\end{algorithm}

The pseudocode is summarized in Algorithm \ref{alg:m3p2i}. After the initialization, we sample Halton splines and forward simulate the plan alternatives using IsaacGym to compute the costs (Lines \ref{line:plan}-\ref{line:end_plan}). The costs are then used to update the weights for each plan and update their means (Lines \ref{line:weights 0}-\ref{line:weights 1}). Finally, the mean of the overall action sequence is updated (Line \ref{line:mean}), and the first action from the mean is executed. 

\subsection{Plan interface}
The plan interface is a component that takes the possible alternative symbolic actions in $\mathcal{P}$ and links them to their corresponding cost functions, forwarding the latter to M3P2I. For every symbolic action a robot can perform, we store a cost function in a database that we can query at runtime, bridging the output of the action planner to the motion planner.
\section{Experiments}
\label{sec:experiments} 
We evaluate the performance of our method in two different scenarios. The first is a \textit{push-pull scenario} for non-prehensile manipulation of an object with an omnidirectional robot. The second is a \textit{object stacking scenario} with a 7-DOF manipulator with dynamic obstacles and external disturbances at runtime.  
\subsection{Push-pull scenario}
\begin{figure}[htb!]
    \centering
    \includegraphics[width=0.24\textwidth]{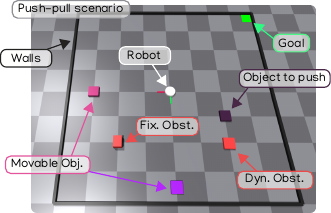}
    \caption{Push-pull scenario. The dark purple object has to be placed on the green area. The robot can pull or push the object while avoiding dynamic and fixed obstacles. The objects and goals can have different initial positions.}
    \label{fig:push-pull_scenario}
\end{figure}
This scenario is depicted in \Cref{fig:push-pull_scenario}. One object has to be placed to a goal, situated in one of the corners of an arena. The object can have different initial locations, for instance, in the middle of the arena or on one of the corners. There are also static and dynamic obstacles, and the robot can push or pull the object. We define the following action templates for AIP and the cost functions for M3P2I. 
\subsubsection{Action templates for AIP}
The AIP for this task requires one state $s^{(goal)}$ and a relative symbolic observation $o^{(goal)}$ that indicates when an object is at the goal. This is defined as:
\begin{align}
    o^{(goal))} = \left\{
    \begin{aligned}
        & 0, ||{p}_{G} - {p}_{O}||\leq \delta \\
        & 1, ||{p}_{G} - {p}_{O}|| > \delta
    \end{aligned} 
    \right. 
\end{align}
\textcolor{black}{where ${p}_{G}, {p}_{O}$ represent the positions of the goal and the object in a 3D coordinate system.} $\delta$ is a constant threshold determined by the user. The mobile robot can either push, pull, or move. These skills are encoded in the action planner as follows:

\begin{table}[h]
\centering 
\begin{tabular}{l l l} 
    \textbf{Actions}& \textbf{Preconditions}& \textbf{Postconditions}\\ 
    \texttt{push(obj,goal)} & \texttt{-} & ${l^{(goal)}} = [1\ 0]^\top$\\
    \texttt{pull(obj,goal)} & \texttt{-} & ${l^{(goal)}} = [1\ 0]^\top$\\\end{tabular}
\end{table}

The postcondition of the action \texttt{push(obj, goal)} is that the object is at the goal, similarly for the pull action. Note that we do not add complex heuristics to encode the geometric relations in the task planner to determine when to push or pull; instead, we will exploit parallel sampling in the motion planner later. \textcolor{blue}{The desired state $s_d$ of} this task is set as a preference for $l^{(goal)} = [1\ 0]^\top$. The BT would contain more desired states for pushing or pulling several blocks. Our approach can be extended to multiple objects in different locations, for instance, and accommodate more involved pre-post conditions and fallbacks since it has the same properties as in \cite{pezzato2022aipbt}. 

\subsubsection{Cost functions for M3P2I}

We need to specify a cost for each symbolic action. 
The cost function for pushing object $O$ to the goal $G$ is defined as:
\begin{equation}
\begin{split}
\label{eq: constraint push}
    C_{push}(R, O, G) &= C_{dist}(R, O)+C_{dist}(O, G)+C_{ori}(O, G) \\
     & +C_{align\_push}(R, O, G)
\end{split}
\end{equation}
where minimizing $C_{dist}(O, G) = \omega_{dist} \cdot ||{p}_{G} - {p}_{O}||$ makes the object $O$ close to the goal $G$. $C_{ori}(O, G) = \omega_{ori} \cdot  \phi(\Sigma_{O}, \Sigma_{G})$ defines the orientation cost between the object $O$ and goal $G$. We define $\phi$ for symmetric objects as:
\begin{equation}
\label{eq: ori_metric}
    \phi(\Sigma_u, \Sigma_v) = \min_{i, j \in \{1, 2, 3\}} \left(2 - ||\Vec{u}_1 \cdot \Vec{v}_i|| - ||\Vec{u}_2 \cdot \Vec{v}_j|| \right)
\end{equation}
where $\Sigma_u = \{\Vec{u}_1, \Vec{u}_2, \Vec{u}_3 \}, \Sigma_v = \{\Vec{v}_1, \Vec{v}_2, \Vec{v}_3 \}$ form the orthogonal bases of two coordinates systems. Minimizing this cost makes two axes in the coordinate systems of the object and goal coincide. \textcolor{blue}{The orientation cost for asymmetric objects can be extended from \Cref{eq: ori_metric} by aligning the corresponding axes.}

The align cost $C_{align\_push}(R, O, G)$ is defined as: 
\begin{align}
\label{eq: constraint align push}
    C_{align\_push}(R, O, G) &= \omega_{align\_push} \cdot h(cos(\theta)),\\
    cos(\theta) &= \frac{({p}_R-{p}_{O}) \cdot ({p}_G-{p}_{O})}{||{p}_R-{p}_{O}|| \cdot ||{p}_G-{p}_{O})||}, \\
    h(cos(\theta)) &= \left\{
    \begin{aligned}
        & 0,\ cos(\theta)\leq0 \\
        & cos(\theta),\ cos(\theta) > 0
    \end{aligned} 
    \right.
\end{align}

\begin{figure}[htb!]
    \centering
    \includegraphics[width=0.24\textwidth]{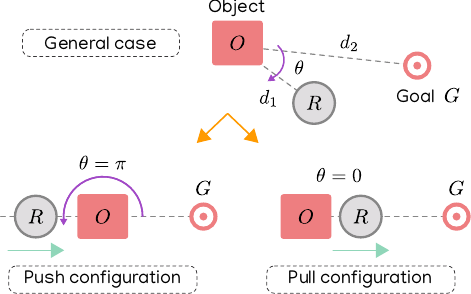}
    \caption{Push and pull ideal configurations. The robot $R$ has to push or pull the object $O$ to the goal $G$.}
    \label{fig:pull_illus}
\end{figure}

This makes the object $O$ lie at the center of robot $R$ and goal $G$ so that the robot can push it, as illustrated in \Cref{fig:pull_illus}.

Similarly, the cost function of making the robot $R$ pull object $O$ to the goal $G$ can be formulated as:

\begin{equation}
\begin{split}
\label{eq: constraint pull}
    & C_{pull}(R, O, G) = C_{dist}(R, O)+C_{dist}(O, G)+C_{ori}(O, G) \\
     & +C_{align\_pull}(R, O, G) + C_{act\_pull}(R, O, G)
\end{split}
\end{equation}
where the align cost $C_{align\_pull}(R, O, G)$ makes the robot $R$ lie between the object $O$ and goal $G$, see \Cref{fig:pull_illus}. While pulling, we simulate a suction force in IsaacGym, and we are only allowed to sample control inputs that move away from the object through $C_{act\_pull}(R, O, G)$. Mathematically:
\begin{align}
    \label{eq: constraint align pull}
    C_{align\_pull}(R, O, G) &= \omega_{align\_pull}\ \cdot h(-cos(\theta)) \\
    C_{act\_pull}(R, O, G) &= \omega_{act\_pull}\ \cdot h(\frac{({p}_O-{p}_R)\cdot \Vec{u}}{||{p}_O-{p}_R|| \cdot ||\Vec{u}||})
\end{align}

\begin{figure}[!htb]
    \centering    \includegraphics[width=0.4\textwidth]{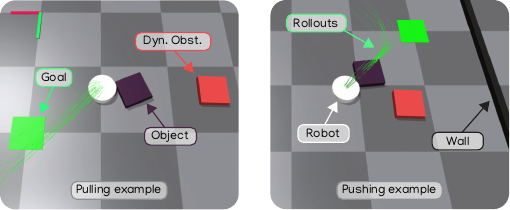}
    \caption{Illustrative example of pulling and pushing a block to a goal. The strategy differs according to the object, goal location, and dynamic obstacle position. What action to perform is decided at runtime through multi-modal sampling.}
    \label{fig:example_push_pull}
\end{figure}

An example can be seen in \Cref{fig:example_push_pull}. \textcolor{black}{We also consider an additional cost $C_{dyn\_obs}(R, D)$ to avoid collisions with (dynamic) obstacles while operating. \textcolor{blue}{The dynamic obstacle is assumed to move in a certain direction with constant velocity.} We use a constant velocity model to predict the position of the dynamic obstacle $D$ in the coming horizon and try to maximize the distance between the latter and the robot:}
\begin{align}
    C_{dyn\_obs}(R, D) = \omega_{dyn\_obs} \cdot e ^{-|| {p}_R - {p}_{D_{pred}}||}
\end{align}
where ${p}_{D_{pred}}$ is the predicted position of dynamic obstacle.
\subsubsection{Results}
We test the performance of our approach in two configurations: a) the object is in the middle of the arena, and the goal is to one corner, and b) both the object and the goals are in different corners. For each arena configuration, we test three cases: the robot can either only push, only pull, or combine the two through our M3P2I. The AIP plans for the two alternatives, pushing and pulling, and forwards the solution to the plan interface. Then, M3P2I starts minimizing the costs until the AIP observes the completion of the task.
We performed 20 trials per case, per arena configuration, for a total of 120 simulations. By only pulling an object, the robot cannot tightly place it on top of the goal in the corner; on the other hand, by only pushing, the robot cannot retrieve the object from the corner. 
Using multi-modal motion, we can complete the task in every tested configuration. \Cref{tab:results_push_pull} shows that the multi-modal case outperforms push and pull in both arena configurations. It presents lower position and orientation errors and a shorter \textcolor{blue}{planning and execution} time. 


\vspace{-5mm}
\begin{table}[htb!]
\caption{Simulation Results of Push and Pull}
\label{tab:results_push_pull}
\centering
\begin{tabular}{ccccc}
    \hline\hline
    \textbf{Case} & \textbf{Skill} & \makecell{\textbf{Mean(std)}\\\textbf{ pos error}} & \makecell{\textbf{Mean(std)} \\\textbf{ori error}} &  \makecell{\textbf{Mean(std) }\\\textbf{time (s)}}\\
    \midrule
     & Push & \makecell{0.1061 \\ (0.0212)} & \makecell{0.0198  \\ (0.0217)} & \makecell{6.2058  \\ (6.8084)} \\
     \makecell{Middle-\\ corner} & Pull & \makecell{0.1898 \\ (0.0836)} & \makecell{0.0777 \\ (0.1294)} & \makecell{25.1032 \\ (13.7952)} \\
     & \makecell{Multi-\\ modal} & \makecell{\textbf{0.1052} \\ (0.0310)} & \makecell{\textbf{0.0041} \\ (0.0045)} & \makecell{\textbf{3.7768} \\ (0.8239)} \\
     \midrule
     & Push & \makecell{7.2679 \\ (3.2987)} & \makecell{0.0311  \\ (0.0929)} & \makecell{time-out} \\
     \makecell{Corner-\\ corner} & Pull & \makecell{0.3065 \\ (0.1778)} & \makecell{0.1925 \\ (0.2050)} & \makecell{32.8838 \\ (7.9240)} \\
     &\makecell{Multi-\\ modal} & \makecell{\textbf{0.1375} \\ (0.0091)} & \makecell{\textbf{0.0209} \\ (0.0227)} & \makecell{\textbf{9.9473} \\ (3.4591)} \\
    \bottomrule
\end{tabular}
\end{table}
\vspace{-5mm}
\subsection{Object stacking scenario}

\begin{figure}[htb!]
    \centering
    \includegraphics[width=0.4\textwidth]{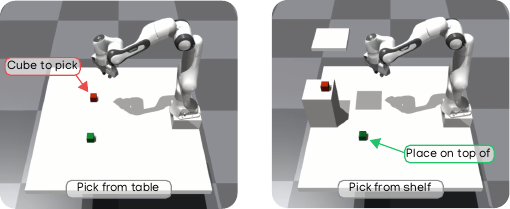}
    \caption{Pick-place scenarios. The red cube has to be placed on top of the green cube. The red cube can be either on the table or a constrained shelf, requiring different pick strategies from the top or the side, respectively.}
    \label{fig:pick_scenario}
\end{figure}

We address the challenge of stacking objects with external task disruptions, necessitating adaptive actions like re-grasping with different pick configurations (e.g., top or side picking in \Cref{fig:pick_scenario}). We showcase the robot's ability to rectify plans by repeating actions or compensating for unplanned occurrences, such as unexpected obstacles obstructing the path. We benchmark against the cube-stacking task outlined in \cite{makoviychuk2021isaac}.

\subsubsection{Action templates for AIP}
For this task, we define the following states $s^{(reach)}$, $s^{(hold)}$, $s^{(preplace)}$, $s^{(placed)}$, and their corresponding symbolic observations. The robot has four symbolic actions, summarized \textcolor{blue}{below:}
\vspace{-2mm}
\begin{table}[!ht]
\label{tab::prepost}
\centering 
\begin{tabular}{l l l} 
    \textbf{Actions}& \textbf{Preconditions}& \textbf{Postconditions}\\ 
    \texttt{reach(obj)} & \texttt{-} & ${l^{(reach)}} = [1\ 0]^\top$\\
    \texttt{pick(obj)} & \texttt{reachable(obj)} & ${l^{(hold)}} = [1\ 0]^\top$\\
    \texttt{prePlace(obj)} & \texttt{holding(obj)} & ${l^{(preplace)}} = [1\ 0]^\top$\\   \texttt{place(obj)} & \texttt{atPreplace(obj)} & ${l^{(placed)}} = [1\ 0]^\top$\\
\end{tabular}
\end{table}

The symbolic observers to estimate the states are defined as follows. To estimate whether the gripper is close enough to the cube, we define the relative observation $o^{(reach)}$. We set $o^{(reach)} = 0$ if $\delta_r \leq \textcolor{blue}{\delta}$, where $\delta_r = ||{p}_{ee} - {p}_{O}||$ measures the distance between the end effector $ee$ and the object $O$. $o^{(reach)} = 1$ otherwise. To estimate whether the robot is holding the cube \textcolor{blue}{of size 0.06m}, we define:
\begin{align}
    & o^{(hold)} = \left\{
    \begin{aligned}
        & 0, \delta_f < \textcolor{blue}{0.06+\delta} \text{ and } \delta_f \geq \textcolor{blue}{0.06-\delta}\\
        & 1, \delta_f \geq \textcolor{blue}{0.06+\delta} \text{ or } \delta_f \leq \textcolor{blue}{0.06-\delta}
    \end{aligned} 
    \right.
\end{align}
where $\delta_f = ||{p}_{ee\_l} - {p}_{ee\_r}||$ measures the distance between the two gripper's fingers. To estimate whether the cube reaches the pre-place location, we define:
\begin{align}
\label{eq:preplace_obs}
    & o^{(preplace)} = \left\{
    \begin{aligned}
        & 0, C_{dist}(O, P) < \textcolor{blue}{\delta} \text{ and } C_{ori}(O, P) < \textcolor{blue}{\delta} \\
        & 1, C_{dist}(O, P) \geq \textcolor{blue}{\delta} \text{ or } C_{ori}(O, P) \geq \textcolor{blue}{\delta}
    \end{aligned} 
    \right.
\end{align}
where $C_{dist}(O, P)$ and $C_{ori}(O, P)$ measure the distance and the orientation between the object $O$ and the pre-place location $P$ as in \Cref{eq: constraint push}. The pre-place location is a few centimeters higher than the target cube location, directly on top of the green cube. We use the same logic as \Cref{eq:preplace_obs} for $o^{(placed)}$ where the place location is directly on top of the cube location. \textcolor{blue}{The desired state for this task is set to be $l^{(placed)} = [1\ 0]^T$}, meaning the cube is correctly placed on top of the other. Note that in more complex scenarios, such as rearranging many cubes, the BT can guide the AIP as demonstrated in \cite{pezzato2022aipbt}. 


\subsubsection{Cost functions for M3P2I}
At the motion planning level, the cost functions for the four actions are formulated as:
\begin{equation}
    \label{eq: constraint reach}
    \begin{aligned}
        C_{reach}(ee, O, \psi) &= \omega_{reach} \cdot ||{p}_{ee}-{p}_{O}|| \\
        &+ \omega_{tilt} \cdot \left(\frac{||\Vec{z}_{ee} \cdot \Vec{z}_{O}||}{||\Vec{z}_{ee}|| \cdot ||\Vec{z}_{O}||} - \psi \right)
    \end{aligned}
\end{equation}
\begin{align}
    &
    \begin{aligned}
        C_{pick}(ee) &= \omega_{gripper} \cdot l_{gripper} 
    \end{aligned}\\
    &
    \begin{aligned}
        C_{preplace}(O, P) &= C_{dist}(O, P) + C_{ori}(O, P)
    \end{aligned}\\
    &
    \begin{aligned}
        C_{place}(O, P) &= \omega_{gripper} \cdot (1-l_{gripper})
    \end{aligned}
\end{align}
$C_{reach}(ee, O, \psi)$ moves the end effector close to the object with a grasping tilt constraint $\psi$. As $\psi$ approaches 1, the gripper becomes perpendicular to the object; as it nears 0, the gripper aligns parallel to the object's supporting plane.

\subsubsection{Results - reactive pick and place}
We first consider the pick-and-place under disturbances. We model disturbances by changing the position of the cubes at any time. We compare the performance of our method with the off-the-shelf RL method \cite{makoviychuk2021isaac}. This is a readily available Actor-Critic RL example from IsaacGym, which considers the same tabletop configuration and robot arm. We compare the methods in a \textit{vanilla} task without disturbances and a \textit{reactive} task with disturbances. It should be noticed that the cube-stacking task in \cite{makoviychuk2021isaac} only considers moving the cube on top of the other cube while neglecting the action of opening the gripper and releasing the cube. In contrast, our method exhibits fluent transitions between pick and place and shows robustness to interferences \textcolor{blue}{such as repick }during the long-horizon task execution. Results are available in \Cref{tab: results reactive pick place}, with 50 trials per case. While the RL agent shows a slightly lower position error in the vanilla case, our method outperforms it in the reactive task. \textcolor{blue}{Planning and execution time for smooth pick-and-place with our method is approximately 5 to 10s.}
\vspace{-5mm}
\begin{table}[!htb]
\caption{Simulation Results of Reactive Pick and Place}
\centering
\begin{tabular}{cccc}
    \hline\hline
    \textbf{Task} & \textbf{Method} & \makecell{\textbf{Training}\\ \textbf{epochs}} & \makecell{\textbf{Mean(std) }\\ \textbf{pos error}}\\
    \midrule
      \multirow{2}{*}{\textcolor{black}{Vanilla}}  & Ours & 0 & 0.0075 (0.0036)\\
      & RL & 1500 & \textbf{0.0042} (0.0019)\\
     \midrule
     \multirow{2}{*}{Reactive} & Ours & 0 & \textbf{0.0117} (0.0166)\\
      & RL & 1500 & 0.0246 (0.0960)\\
    \bottomrule
\end{tabular}
\label{tab: results reactive pick place}
\end{table}
\subsubsection{Results - multi-modal grasping} 
\begin{figure}[htb!]
    \centering
    \includegraphics[width=0.4\textwidth]{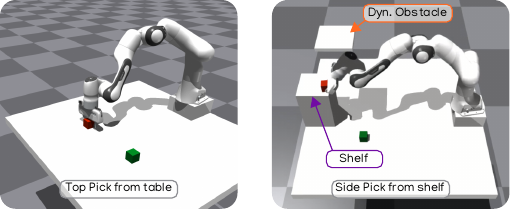}
    \caption{Example of different picking strategies computed by our multi-modal MPPI. The obstacle on top of the shelf can be moved, simulating a dynamic obstacle.}
    \label{fig: generalizing pick}
\end{figure}

In this case, we consider grasping the object with different grasping poses by sampling two alternatives in parallel. That is, pick from the top or the side to cover the cases when the object is on the table or the constrained shelf with an obstacle above. To do so, we use the proposed M3P2I and incorporate the cost functions of $C_{reach}(ee, O, \psi=0)$ and $C_{reach}(ee, O, \psi=1)$ as shown in \Cref{eq: constraint reach}. This allows for a smooth transition between top and side grasp according to the geometry of the problem, see \Cref{fig: generalizing pick}. 

\subsubsection{Results - real-world experiments}
\begin{figure}[htb]
    \centering
    \includegraphics[width=0.4\textwidth]{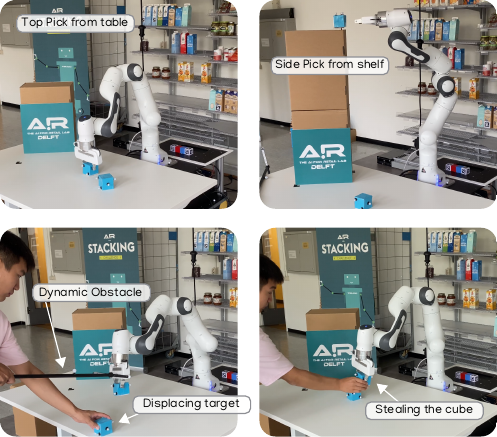}
    \caption{Real-world experiments of picking a cube from the table or the shelf while avoiding dynamic obstacles and recovering from task disturbances.}
    \label{fig: real-world pick}
\end{figure}

\textcolor{black}{Our real-world validation of reactive pick-and-place, depicted in \Cref{fig: real-world pick}, involves avoiding a moving stick and disturbances such as movement and theft of the cube. M3P2I enables smooth execution and recovery while using different grasp configurations.}

\section{Discussion}



\textcolor{blue}{In this section, we discuss key aspects of our solution and potential future work. 
The main strength of M3P2I is its ability to reason over discrete alternative actions at the motion planning level. This is enabled by sampling different control sequences for each alternative symbolic action and then blending them through importance sampling. We thus alleviate the task planning burden by eliminating logic heuristics to switch between these actions. Sampling alternatives at the motion planning level increases robustness during execution, at the price of slightly degrading the performance since the control distribution is also slightly biased towards less effective strategies, as shown in \cite{trevisan2024biased}.
The performance of M3P2I also depends on the weight tuning of the cost functions. In this case, implementing auto-tuning techniques can reduce manual effort \cite{spahn2023autotuning}. The cost functions also need to capture the essence of the skills.}
\textcolor{blue}{
The AIP requires manually defined symbolic action templates and a set of discrete states. The discrete desired states need to be encoded in a sequence in a BT or can be as simple as encoding the end state for a task, as in our examples.}
\textcolor{blue}{To transfer from simulation to the real world, we considered randomization of object properties in the rollouts~\cite{pezzato2023sampling}. Online system identification could be added to achieve better performance with uncertain model parameters~\cite{abraham2020model}.}

\section{Conclusion}
\label{sec:conclusions}
In this paper, \textcolor{black}{to address the runtime geometric uncertainties and disturbances}, we proposed a method to combine the adaptability of an Active Inference planner (AIP) for high-level action selection with a novel Multi-Modal Model Predictive Path Integral Controller (M3P2I) for low-level control. We modified the AIP to generate plan alternatives that are linked to costs for M3P2I. The motion planner can sample the plan alternatives in parallel, and it computes the control input for the robot by smoothly blending different strategies. In a push-pull task, we demonstrated how our proposed framework can blend both push and pull actions, allowing it to deal with corner cases where approaches only using a single plan fail. With a simulated manipulator, we showed our method outperforming a reinforcement learning baseline when the environment is disturbed while requiring no training. Simulated and real-world experiments demonstrated how our approach solves reactive object stacking tasks with a manipulator subject to severe disturbances and various scene configurations that require different grasp strategies.


\bibliographystyle{IEEEtran}
\bibliography{IEEEabrv,mybib}

\end{document}